\documentclass{article}
\usepackage{graphicx}
\usepackage{tabularx}
\usepackage{tabulary}
\usepackage{fullpage}
\usepackage{etoolbox}
\usepackage{multirow}

\makeatletter
\patchcmd{\@makecaption}
  {\scshape}
  {}
  {}
  {}
\makeatletter
\patchcmd{\@makecaption}
  {\\}
  {.\ }
  {}
  {}
\makeatother

\begin{document}

\title{Social Robot Modelling of Human Affective State}

\author{D.B. Skillicorn, N. Alsadhan, R. Billingsley and M.-A.- Williams}

\maketitle

\begin{abstract}
Social robots need to understand the affective state of the humans
with whom they interact. Successful interactions require understanding
mood and emotion in the short term, and personality and attitudes
over longer periods. Social robots should also be able to infer
the desires, wishes, and preferences of humans without being
explicitly told.

We investigate how effectively affective state can be inferred from
corpora in which documents are plausible surrogates for
what a robot might hear.
For mood, emotions, wishes, desires, and attitudes we show highly
ranked documents; for personality dimensions, estimates of
ground truth are available and we report performance accuracy. 
The results are surprisingly strong given the limited information
in short documents.
\end{abstract}

\section{Introduction}

Social robots must build and use models of the humans with whom they interact.
The quality of these interactions is much improved if
robots are able to estimate the affective state of such humans.

Affective state can be considered as being made up of the following
components, about which there is wide agreement among 
psychologists \cite{has}:
\begin{itemize}
\item
Mood.
This captures a basic state of mind that is internally generated
(although the drivers are not well understood) and relatively long
lasting. It can be divided into two components: positivity and
negativity, which are independent rather than opposites. In other
words, an individual can have increased positivity, without necessarily
having reduced negativity. Mood acts to dampen the intensity of 
countervailing emotions.
\item
Emotions.
Emotions are a parallel channel to cognition that focuses on the
values associated with external situations. Studies of humans with
brain injuries that reduce emotional intensity indicate that such
injuries cause difficulty making decisions, suggesting that a critical
part of the function of emotions is to rank the importance
or significance associated with potential goals and tasks.
Emotions are primarily a reaction to external circumstances.
There are a number of categorization of basic emotions, but we
will use anger, disgust, fear, and sadness (which are considered negatively
associated emotions); and anticipation, joy, and surprise (which
are considered positively associated emotions).
High positive mood tends to enhance the expression of positive
emotions and dampen the expression of negative emotions, and
the converse.
\item
Personality.
One definition of personality is ``stable differences in motivational
reactions to circumscribed classes of environmental stimuli" \cite{denissen}. 
In other words, personality captures the toolkit of reactions (often
called ``reaction norms") that each individual preferentially uses.
Personality is often quantified in terms of the so-called
Big Five dimensions, with acronym OCEAN: openness, conscientiousness, 
extraversion, agreeableness,
and neuroticism. Each dimension actually represents an opposite
pair, for example extroversion-introversion, and an individual is placed 
on axes corresponding to their expression of each of these five property
pairs. 
It is typically assumed that personality is
fixed (at least by adulthood) for each individual.
\item
Values.
These associate positivity or negativity with both physical and
abstract objects: concepts, beliefs, and
goals. They are primarily cognitive.
\item
Attitudes.
These are evaluations of specific instances of people, ideas,
events, situations, and other more-concrete objects, providing
shorthand guidelines for the proper responses to them, including
potential actions.
In an ideal world, attitudes would align with, and so be predictable
from, values but few individuals exhibit this level of consistency.
\end{itemize}

\begin{figure}[h]
\centering
\includegraphics[width=0.45\textwidth]{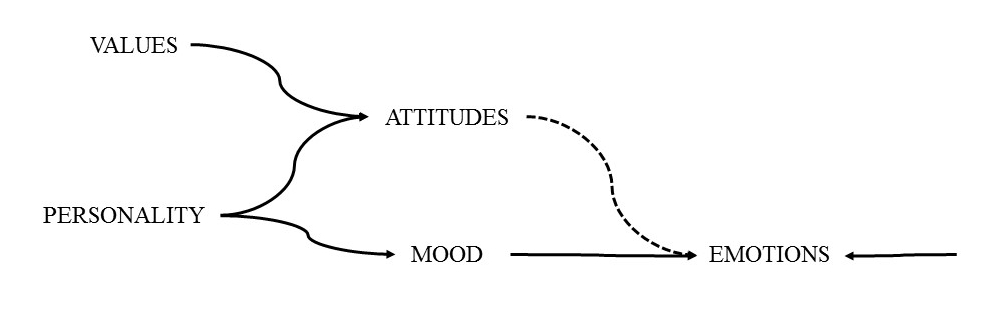}
\caption{Relationship among components of affective state}
\label{affective}
\end{figure}

The relationships between these components of affective state are
shown in Figure~\ref{affective}.
Components on the left are relatively fixed; they influence
components further to the right, which are more volatile: values and
personality fixed over a major fraction of a lifetime, attitudes fixed
over years, mood fixed over a day, and 
emotions varying on a time scale of seconds.

Why should a robot take into account the affective state of
a human with whom it is interacting? 
Social interactions assume an understanding of affect by all parties,
and some autism-spectrum disorder individuals have difficulty in
social settings precisely because they lack such understanding.
A robot cannot be called `social' without at least some awareness
of the role of affect.
The feedback loop that seems to be required to create effective
interactions, between human and human and now between human and robot,
has been called the \emph{affective loop} \cite{affectiveloop}.

In particular, the ability to infer a human's emotions is useful
for the following reasons:
\begin{itemize}
\item
The robot can compute more accurately the priorities that the
human will have for his/her own behavior.
\item
The robot can improve its own prediction of the desires of
the human, and so can make itself more useful.
\item
The robot can detect when it is performing poorly from 
the human's perspective.
\item
The robot can mimic the human more effectively to create better rapport.
\item
The robot can act to help alter the human's emotional state, when appropriate.
\end{itemize}

Because emotions change rapidly, a robot's model of a particular
human's emotion must update frequently.
The process for inferring emotion must therefore be lightweight.
Since emotions are volatile, they can only be inferred from recent
data about each human, and consequently from small amounts of data.

The reason to infer a human's mood is that inference of
emotions cannot work well without modelling the effect that
mood has on them.
We might casually say that `X is an angry person' but it would
be more accurate to say that `X's is habitually in a strongly
negative mood, and this amplifies his emotion of anger'.

It might be imagined that another motivation for a robot
to infer a human's mood might be to try and alter it.
When someone is in a `bad mood' we often talk about trying to
cheer them up. However, technically, moods are only very weakly
influenced by others or the environment. What we mean when we
talk of cheering someone up is really to act to reduce their emotion 
of sadness.
The process of altering another's emotional state deliberately
is quite a difficult one, even for humans; it may eventually be
possible to teach robots to do this, but it is beyond our present
capabilities.

The reason to infer the human's personality is that it delimits
the repertoire of normal reactions that each individual human will choose from
in any particular situation, and therefore allows the robot to
anticipate more accurately. Different humans behave differently,
even when they are in, say, the same bad mood and feeling angry; and
a social robot must be able to interact with each human as an individual.

Because personality is relatively fixed, a robot can infer a particular
human's personality based on the totality of data about that human,
perhaps refining it incrementally. More complex models can be built,
since they can be developed and improved as a background process.

Computational approaches to values and attitudes are not well developed
except in contexts such as marketing and advertising, where they
are often further narrowed to attitudes towards particular products.
The difficulty with modelling attitudes is that they are really
relations between an individual and external objects, rather
than an internal property of the individual.
The ability to infer values and attitudes will be useful to a 
robot in determining a human's responses to the current context, 
and immediate intentions. We explore the use of the semantic
differential as a first step towards computational modelling of
attitudes.

\section{Modelling Strategy}

We must use data that resembles what a robot might (over)hear in the
presence of a human whom it is trying to model, and analysis
that can plausibly be carried out in an environment
with limited time and resources. 
For example, full parsing, or even part-of-speech tagging, may
be too expensive, and perhaps not effective given the fragmentary,
non-grammatical nature of informal speech.
For those properties for which models must be built quickly, from
small amounts of data, accuracy cannot be expected to be high.
A human might be able to infer the emotional state and mood of
another human from a few utterances, but would typically need 
more to infer personality.

Approaching affective state lexically has been a popular
approach for almost a century. As a result, a number of lexicons
have been developed. In each case, word frequencies are considered
to be signals of a component of affective state. Empirical studies
have shown that even emotional state -- for which more than 500
distinguishable emotions have been suggested -- can be described
using a much smaller set of core or underlying emotions, and
this is characteristic of other aspects of affective state.

Components can be unipolar or bipolar. For example, the
emotion of anger is unipolar: one can be angry, but the
opposite is a neutral state of not being angry. 
In this case, a count of words from
a lexicon of anger-related words estimates how angry the speaker
or writer of an utterance is.
Emotions are usually regarded as unipolar from a psychological
perspective (although in common parlance, many emotions have named
opposites, e.g. angry-calm).

The Big Five personality traits are usually conceived of as
bipolar; openness is really a pair: openness to experience versus
closedness to experience. A lexicon for bipolar components
differentiates words based on which `direction' they are associated
with.

Our basic approach is to use lexicons associated with each
property of affective state we are interested in. These are:
\begin{itemize}
\item
For mood, positive and negative lexicons from the NRC Emotion
Lexicon \cite{mohammad:nrc}.
\item
For emotions, the seven lexicons for anger, disgust, fear, sadness,
anticipation, joy, and surprise from the NRC emotion lexicon
\cite{mohammad:nrc}. Plutchik \cite{plutchik} listed these
as primary emotions (including trust, which seems to be an attitude
rather than an emotion, that is a relation rather than a state).
There are many overlapping sets of `basic' emotions, many of
them deriving from the circumplex model of Russell \cite{russell:circumplex}.
For a more physiological discussion of basic emotions, see
Ekman \cite{ekman:emotions}.
\item
A lexicon derived from the Wishes dataset to see whether it is
plausible for a robot to detect an expression of a wish, that is
an implicit command \cite{wishes}.
\item
A lexicon, developed by the authors, for preferences 
to see whether it is plausible for a robot
to detect desires/preferences in humans.
\item
A lexicon based on the Osgood scores of the most common 1000 words
in English, capturing the dimensions of good-bad, active-passive,
and strong-weak \cite{osgood}. 
These have been shown to underlie attitudes across many cultures.
\end{itemize}
The relationship between personality and word usage has mostly been
explored from the perspective of how personality influences
words \cite{schwartz,lee} and the related efforts to understand
the underlying dimensions of variation in word usage (that is, to
justify the Big Five -- sometimes Big Six -- dimensions).
Attempts to predict personality from word usage (for example,
\cite{farnadi2013recognising,farnadi:sitaraman}) have used varying 
approaches to what is being
predicted (one dimension at a time or an entire personality) and
what attributes are used, often adding syntactic attributes such as
n-grams to lexical attributes.

Others have studied the converse problem, methods for robots to
express emotion in ways that seem natural to humans (thus closing the
other side of the affective loop) \cite{kowalczuk,kirby}.

Attitudes have received less attention, except for the vast literature
on sentiment analysis -- the attitudes of individuals to particular
products. More general approaches include Liu and Maes attempt
to build a model of the attitudes of other humans that could be used
as an advice surrogate \cite{liu:maes} and more linguistic work
on the mechanisms by which attitude differences are expressed in
language \cite{polyani}.

\section{Experiments}

We use the following datasets in our experiments:
\begin{itemize}
\item
A set of essays written by university 
students about their daily lives and thoughts, 
collected by Pennebaker between 1997 and 2004 and labelled
with personality classes \cite{pennebaker1999linguistic}.
\item
A set of Facebook posts collected by Stillwell and Kosinski using a 
Facebook application that implemented the Big Five test using 
Costa and McCrae's NEO-PI-R domains, along with
other psychological tests \cite{costa1995domains}.
\item
A set of transcripts of video blogs (vlogs) from Youtube, labelled with
personality impressions by crowdsourcing \cite{biel2013hi}.
\item
The first 100,000 tweets in the Microsoft Research Conversation Corpus,
a set of 4.46 million tweets from about 1.3 million conversations.
\item
A set of 1250 headlines from the BBC, labelled with both mood and
emotions by human raters, and used as a problem dataset for the
Semantic Evaluation workshop in 2007 \cite{semeval2007}.
\end{itemize}
The sizes and properties of these
datasets are shown in Table~\ref{dataprops}.

Facebook posts and tweets (and perhaps headlines)
are good representations of what a robot
might overhear, and can plausibly be used to infer mood and emotions.
The essays and Youtube datasets are less useful because these documents were
produced over longer time frames, during which emotions might change.

Properties such as mood and personality are more stable over time, so
the essays and Youtube datasets are more useful for estimating them.

\begin{table}[]
\def\arraystretch{1.1}%
\centering
\caption{Dataset properties}
\label{dataprops}
\begin{tabular}{|l|l|l|}
\hline
	        & \textbf{Number of} & \textbf{Av. doc. length}  \\
\textbf{Dataset} & \textbf{documents} & \textbf{(\& with rare words}  \\
	       &  & \textbf{removed)} \\ 
\hline
\textbf{Essays}         & 2469          & 577 (563.5) \\ 
\hline
\textbf{Facebook posts} & 9918          & 10.9 (8.6) \\ 
\hline
\textbf{Tweets}    & 100,000 & 8.99 (8.99) \\ 
\hline 
\textbf{Headlines}  & 1250 &  6.5 ($<1$)) \\
\hline
\textbf{Video blogs} & 404 & 587 (480) \\
\hline
\end{tabular}%
\end{table}

For each of these datasets and each lexicon, we build a document-word
matrix describing the counts of the words in each document.

Variations in document length are an issue for all corpora,
because of their effect of document-document similarity.
In the essays corpus, documents are of similar 
lengths;
the Facebook posts vary in length from a single word to
several hundred words;
tweets have an imposed upper bound of 140 characters, but we observe a
non-trivial number of very short tweets;
headlines are all very short;
and the Youtube vlogs are longer but have significant length variation.

When normalizing to take document lengths into account,
we address two underlying processes.
One is the occurrence of repeated words, which are far more
common in tweets and Facebook posts than they would be in ordinary
speech, apparently as a mechanism for emphasis.
To compensate, we take the logarithms of word counts.
This, for instance, reduces the count of positive or joyful words 
in a document like "I'm happy, happy, happy" in comparison to 
"I'm happy, I'm ecstatic", in which the expression of emotion
has required more cognitive processing.

The second is that the average word counts of longer documents are
inevitably longer than the word counts of shorter documents.
The conventional way to address this is to divide each word count by
the length of the document in which it appears. This works reasonably
well for documents of modest length but, for short documents, the
denominator is small and the resulting word rates become distortingly
large.
To compensate, we divide word counts in a document by the digamma function 
of the total number of words it contains. 
The digamma function has the shape of a logarithm
but rapidly becomes much flatter. It quite accurately fits the shape of the
graph of new words encountered as a function of words encountered
in ordinary documents \cite{chandra}. Its Taylor series expansion to a few terms
is also quite a good approximation, so it can be computed inexpensively.
Applying this transformation modestly increases the weights
of words that appear in short documents while having little effect on
words that appear in longer documents.

The reader is cautioned that Tweets and Facebook posts are presented 
intact, and so contain inappropriate language.

\subsection{Mood}

Posts are processed to extract word counts for words associated
with positivity and negativity lexicons, these word counts are normalized
for document length and scores are computed by
summing the normalized word weights for each document.
This is, surprisingly, more effective than conventional
techniques such as computing the singular value decomposition of 
the document-word matrices.

The only corpus of which we are aware that is labelled with 
multiple emotions is the headlines corpus. It was labelled by
humans, but with only weak agreement between human raters \cite{semeval2007}.
Attempts to predict the human-generated labels fail because
they clearly made their judgements without using many of the
words of the NRC emotion lexicon. The problem
seems to be not that the NRC lexicon is deficient, but that the human
raters made their judgements using considerable contextual knowledge
rather than the words that actually appear in the headlines.
We provide comparisons of the top-ranked headlines by lexicon
score and by human rater for each emotion.
Both approaches do not fare well (possibly because headlines
are highly stylized text).

For unlabelled data, we show ranked lists of the most highly
ranked documents with respect to the property being considered.
Table~\ref{postweets} shows top-scoring positive mood tweets while
Table~\ref{posfb} shows top-scoring positive mood Facebook
posts.
Scores for tweets are integers because they are derived from word
counts; scores for Facebook posts are non-integers because word
counts have been changed to word pseudo-rates by document
length normalization.
Table~\ref{posheads} shows the top-scoring positive mood headlines
both as ranked by human raters, and using the word
counts from the mood lexicon. The human
scores are given as an average value between $+100$ and $-100$,
from most positive to most negative.

\begin{table*}
\centering
\caption{Top positive mood tweets. The limitations of a purely lexical
approach can be seen in  the 3rd ranked tweet, where words like 'truth'
and 'home' increase the positivity score when, from the context, they should
not. Note that strategies like calculating the 'net positivity' are
not valid because positivity and negativity are independent properties.}
\label{postweets}
\begin{tabularx}{\textwidth}{|l|X|}
\hline
Score & Tweet \\
\hline
7 & kinda... probably got a job... a good job too... a legit honest to god good job... you should ask me what it is :] \#fb \\
7 & RE: High Road - Glad to help. U pray I pray We all pray. God provides. Our God is an awesome God! keep on keepin on \\
7 & user11571 truth hurts buddy. We deserved it in 95 with 20 home and away wins. Good luck with the dance action. \\
7 & rock a by baby on the tree top, when the wind blows the craddle will rock lol! or maybe think good thoughts, like ya beautiful! \\
6 & want to eat some rich chocolate icecream covered with decadent chocolate sauce with whipped cream and a cherry on top and rainbow sprinkles! \\
6 & - morning and happy national day! good nite sleep often lead to a good start for the day. enjoy the rest of your day! \\
6 & It won't be long until you find love, and by love, I mean TRUE love. You are such a beautiful, sweet, intelligent, \\
6 & VERY SWEET of you! Thank you for the compliment. Rock on with Peace, Love, \& Respect Always! \\
\hline
\end{tabularx}
\end{table*}

\begin{table*}
\centering
\caption{Top positive mood Facebook posts. The 7th ranked post is quite
negative, but also contains strong positive elements.}
\label{posfb}
\begin{tabularx}{\textwidth}{|l|X|}
\hline
Score & Post \\
\hline
2.0019 & Let me know you more deeply and truely, oh Lord. There are none who know you completely, and so I pray that YOU show me. Your ways are perfect and I want my life to sing your praise. Let me walk with love, joy, peace, patience, kindness, goodness, gentleness, and self-control. Against such things there is no law.  \\
1.8461 & Hallelujah! Hallelujah! Holy! Holy! Are you Lord God almighty! Worthy is the Lamb, worthy is the Lamb that was slain! Unto you be all glory and honor and praise! May my life bring glory to your name :)  \\
1.6792 & Define me urban dictionary   The Most wonderful person in the World. Kind, Sweet, Loving, Caring, Gentle. Perfect in Every way. The one you love for all your life. Crazy hot girl. Beautiful, smart \& funny; *PROPNAME* posesses atributes absent in 99.9\% of women. Truly a lucky find. Plus she rocks. The most wonderful drug in the world, better know as lortab. taking the pill may cause a sense of euphoria, \& well being.   hahaha  \\
1.4489 & The star burnned like a flame, pointing the way to God, the King of kings; the wise men saw the sign and brought their gifts in homage to their great King.      Seeing the star, the wise men said: This must signify the birth of some great king. Let us search for him and lay treasures at his feet: gold, frankincense, and myrrh.    \\
1.3207 & is wishing a Happy Birthday to my bestie *PROPNAME*!   I love you honey, and I hope you have a wonderful 22nd Birthday!    \\
1.3181 & The Soup: The Verdict Type: Vegetable Appearance: Fantastic Texture: Good Taste: Poor Overall: Edible but not pleasant Summary: This is a great soup for looking at, it is full of colour, healthy vegetables and creaminess! If you are not hungry make this soup and you will feel good! However, if eating is your thing, this soup should be avoided. I have soup left for 3 meals - any takers? Only p\&p to pay! Enjoy!  \\
1.3024 & Please keep *PROPNAME*'s family in your prayers tonight and tomorrow. We have to be in court at 9am. Really pray for the oldest *PROPNAME* went to his dad's for his summer visit and had his mother, brother and sister stripped from him for six months. CPS is providing Christmas for the other two but not him.  I sure hope we succeed at stripping them of their   immunity   and make them pay for what they've done to this family.  \\
1.2794 & Two weddings, one baby shower and a great church service on Sunday! A weekend full of love and encouragement and I know the best is yet to come!   All things work together for the good to them that love God, for them who are the called according to his purpose.    \\
1.2473 & money, teeth cleaning, g-ma's, bus, top secret errand, target, rodarte dress for \$40(!!!), some warm and comfy- themed xmas shopping, spending time with mom, coffee, sephora.com, the office, stepping outside and forgetting how cold it gets without proper coverage- productive and selfindulgent bliss. not a bad day off.  \\
1.2473 & Feeling DAMN GOOD headwise lately: forceful self-change of outlook, meds seem to be finally working, \& I'm startin to get shit done! 'Course, meds need tweaking--one of 'em (dunno which) is also tryin to make me a zombie--not the fun movie kind, either: I get sleepy around 4 or 5 PM now; too damn early even if I weren't a night owl! PSYCHED bout the forward progress, but this zombie thing's crampin' my nightlife, oy!  \\
\hline
\end{tabularx}
\end{table*}

\begin{table*}
\centering
\caption{Top positive mood headlines. The human-selected headlines require
considerable context to understand why they are positive (and the first one
remains mysterious); the lexically ranked headlines struggle with the
extremely short documents.}
\label{posheads}
\begin{tabularx}{\textwidth}{|l|l|X|}
\hline
Method & Score & Post \\
\hline
 & 87 & Goal delight for Sheva \\
Human & 85 & Pamuk wins Nobel Literature prize \\
 & 84 & Amazon.com has 'best ever' sales \\
\hline
 & 3 & Full recovery expected for marathon winner \\
Lexical & 3 & Flu Vaccine Appears Safe for Young Children \\
 & 3 & News Baby pandas! Baby pandas! Baby pandas! \\
\hline
\end{tabularx}
\end{table*}

Table~\ref{negtweets} shows top-scoring negative mood tweets, while
Table~\ref{negfb} shows top-scoring negative mood Facebook posts.
Table~\ref{negheads} shows the most negatively ranked headlines,
again based on both human raters and the lexicon.

\begin{table*}
\centering
\caption{Top negative mood tweets. The top ranked post shows the
limitations of a purely lexical approach, misreading an onomatopoeia
for a negative word. The 8th ranked post is also poorly ranked
because of words like 'lost', 'small, and 'insignificant'. Consider,
though, the amount of context that is needed to know that this is a
post about weight loss (good) rather than currency loss (bad)
in the U.K.}
\label{negtweets}
\begin{tabularx}{\textwidth}{|l|X|}
\hline
Score & Tweet \\
\hline
8 & dun dun dununun dun dun! dun dun dununun dun dun! "Ghostbusters!" \\
7 & Had to tackle some skinny ass white mother fucker! His stupid ass almost made me miss my train. I refuse to wait 15 minutes cause you dumb. \\
7 & Is fed up. Bloody no job. Bloody no money. Bloody hospital. Bloody drugs. Bloody boredom. Bloody buggering bollocks. \\
6 & RT user4193: \#dontyouhate thirsty ass bitches/niggaz ughhh go somehwere with ur thirsty ass!! (Byeeee! Go die in the desert!!!) \\
6 & Might have to skip work tomorrow... sore throat+sneezing+fever= bad news :( no, not swine flu! My Mum had this same thing a few days ago. \\
5 & hell yeah that shit bomb i got everybody n the hood doin that shit \\
5 & perhaps sgt. crowley should personally reimburse mass. tax payers for his stupidity \& wasteful expenditure of tax dollars. \\
5 & I can has lost another pound and a half! That's a quarter of a stone, small but not insignificant amount! \\
5 & tell that bitch she dont want it; ill kick her ass! tryna mess with my steph smh bird! \\
\hline
\end{tabularx}
\end{table*}

\begin{table*}
\centering
\caption{Top negative mood Facebook posts. The 3rd ranked post is interesting
-- the poster is obviously highly positive, but the post contains a large
number of negative words and might arguably be considered to express
considerable negativity as well.}
\label{negfb}
\begin{tabularx}{\textwidth}{|l|X|}
\hline
Score & Post \\
\hline
1.5812 & (continued from above) And rot inside a corpse's shell The foulest stench is in the air The funk of forty thousand years And grizzly ghouls from every tomb Are closing in to seal your doom And though you fight to stay alive Your body starts to shiver For no mere mortal can resist The evil of the thriller (Into maniacal laugh, in deep echo)  \\
1.4481 & Recidivism (n.) - 1. Committing new offenses after being punished for a crime.  2.  Chronic repetition of criminal or other antisocial behavior. Replace   criminal   with   stupid  , and I think I've found my problem.  \\
1.1806 & I'm ecstatic! I just spoke briefly with *PROPNAME*'s lawyer. Based on some   case files   I emailed her, we have enough evidence to prove CPS's removal of the kids was illegal and grounds for an emergency hearing. We believe we have enough evidence to cost both CPS and Humble PD their   qualified immunity  , allowing us to file a civil suit against them and the apartments. ANYONE KNOW A GOOD CIVIL ATTORNEY???  \\
1.1303 & Fight one more round. When your arms are so tired that you can hardly lift your hands to come on guard, fight one more round. When your nose is bleeding and your eyes are black and you are so tired that you wish your opponent would crack you one on the jaw and put you to sleep, fight one more round – remembering that the man who always fights one more round is never beaten  \\
1.1168 & will gladly boast in his weaknesses. When I am weak, then He is strong. Your might, oh Lord is endless. Please cover over me. Please cover over your beloved..Lest we  all fall in overwhelming shame under the weight of our imperfections.  \\
1.0684 & Heh...:  God I wish that I could hide away//And find a wall to bang my brains//I'm living in a fantasy,//a nightmare dream...reality//People ride about all day//In metal boxes made away//I wish that they would drop the bomb//And kill these cunts//that don't belong! I hate people!//I hate the human race//I hate people!//I hate your ugly face//I hate people!//I hate your fucking mess//I hate people!//They hate me!  -Anti-Nowhere League  \\
1.0555 & Like fire, Hellfire,  This fire in my skin.        This burning, desire,         Is turning me to sin.  | It's not my fault; I'm not to blame.   It is the gypsy girl; The witch who sent this flame  | It's not my fault; If in God's plan.  He made the devil so much stronger than a man  \\
1.0539 & I have a ship and a fighting crew and a girl with lips like wine, and that's all I ever asked.  Lick your wounds, bullies, and break out a cask of ale.  You're going to work ship as she never was worked before.  Dance and sing while you buckle to it, damn you! To the devil with empty seas! We're bound for waters where the seaports are fat, and the merchant ships are crammed with plunder!    \\
1.0028 & I don't get some people...you offer to help them- they are lazy and don't bother to send their shit... months later when its too damn late they ask you to help them.. wtf \\
0.99347 & Sick of this nihilistic depression shit--fine, life may be meaningless, I may be right about all the other disturbingly bleak shit I've been thinking...but I'm not a self-killer so I'm stuck here--might as well pull it together and live this crazy life I've got anyway, eh?  Some change is in the works (just in time for the New Year, ironically).  \\
\hline
\end{tabularx}
\end{table*}

\begin{table*}
\centering
\caption{Top negative mood headlines. Again the human-ranked headlines
require more context to recognise their negativity.}
\label{negheads}
\begin{tabularx}{\textwidth}{|l|l|X|}
\hline
Method & Score & Post \\
\hline
 & -100 & Bombers kill shoppers \\
Human & -98 & Mortar assault leaves at least 18 dead \\
 & -98 & Iraq car bombings kill 22 People, wound more than 60 \\
\hline
 & 4 & Brazil Air Force Cites Faults and Confusion in Fatal Crash \\
Lexical & 4 & Freed kidnap suspect: my terror at police raid \\
 & 4 & Iraqi suicide attack kills two US troops as militants fight purge \\
\hline
\end{tabularx}
\end{table*}

For the essays dataset there is no need to normalize for document
length, since all of the essays are relatively long, and of similar
lengths. Empirically, normalizing for repeated words seems to improve
the quality of the results slightly. Since positivity and negativity
are not anti-correlated, the same essay can
score highly for both positivity and negativity. Essays are too
long to include in their entirety, but some example
extracts are shown in Table~\ref{moodessays}.

\begin{table*}
\centering
\caption{Extracts from essays scoring high on positivity and/or negativity}
\label{moodessays}
\begin{tabularx}{\textwidth}{|X|l|}
\hline
Post & Comment\\
\hline
I really don't want to go back to LOCNAME, I get lonely. I hate this 
stupid town, but I don't want to leave home, but then again I do , 
the air is cold when I breathe in  and I need to go take my medicine  
oh well I'll go do that later  crap I'm tired and I don't want to 
go read either it gets so long and boring & very negative essay \\
\hline
I feel gross because it's the end of the weekend and I was out partying  the whole time. ... But, I am having so much fun. & essay that is both positive and negative \\
\hline
I started to cry, out of sheer joy at what he was saying to me, I have never had any one person ever express their feelings to me & strongly positive essay \\
\hline
\end{tabularx}
\end{table*}

The video blog transcripts have similar properties. The documents are longer,
and with similar lengths, so we normalize only to discount repetitions.
The highest scoring positive and negative mood documents are shown in
Table~\ref{moodvlogs}.

\begin{table*}
\centering
\caption{Extracts from vlogs scoring high on positivity and negativity}
\label{moodvlogs}
\begin{tabularx}{\textwidth}{|X|}
\hline
Post \\
\hline
Well, I am so excited to talk to you today about goals, strategy and 
tactics, because I -- this is really the secret sauce.
First off, the goal. Might sound like a no brainer but some people forget 
what it is, to increase revenue. Hello, we're in the business of making 
money, that's what we're supposed to be doing here. 
Uh, now a lot of people think of the strategy as being the goal. 
The strategy I'm going to use is to increase audience. 
That's a strategy to achieve that goal. \\
\hline
Okay, this is part two of my anarchy Q and A session. XXXX says I vote 
for capitalism, which, like democracy, is a horribly flawed system that 
happens to be the best thing we can come up with at the moment. 
However, even capitalism is starting to show its flaws like the social ones. 
I seriously doubt capitalism will survive another fifty years at current 
rates of technological development. We're perilously close to technologies 
capable of surpassing the abilities with respect to the specific jobs of 
most of the morons in our population. \\
\hline
\end{tabularx}
\end{table*}

\subsection{Emotions}

Documents are processed in exactly the same way using lexicons for
three positive emotions: anticipation, joy, and surprise; and four
negative emotions: anger, disgust, fear, and sadness.
Tables \ref{anticipationtweets}--\ref{sadnessheads}
show the top scoring tweets, Facebook posts, and headlines for
each emotion (except that anticipation was not included in the
Semeval data).
We appeal to face validation, demonstrating that the highly ranked
examples for each dataset and each emotion do, on the whole,
appear to exhibit that emotion strongly.

\begin{table*}
\centering
\caption{Top anticipation tweets. Note the (undetected) sarcasm in
the 10th ranked tweet.}
\label{anticipationtweets}
\begin{tabularx}{\textwidth}{|l|X|}
\hline
Score & Tweet \\
\hline
7 & RE: High Road - Glad to help. U pray I pray We all pray. God provides. Our God is an awesome God! keep on keepin on \\
6 & good luck good luck good luck! \\
5 & - morning and happy national day! good nite sleep often lead to a good start for the day. enjoy the rest of your day! \\
5 & thought of ya the other day. Time will come and just remember all the good stuff ya can. God bless! \\
5 & hey! just wanted to say ive enjoyed praising god with you at church! You have an amazing gift! Im glad you are getting to share it \\
5 & yes its as long as U wake up alive it's enough reason 2 celebrity life \& enjoying life, we make R self happy, happines is not gained \\
5 & LETS GO GIANTS!! ::clap, clap, clap clap clap!: \\
5 & glad you had a good time, thanks for coming.  Hope to see you at the next one which should be in 2 weeks. \\
5 & sure it's just as good... like white chocolate is just as good as real chocolate. \\
5 & Yh you did start it! Bloody hell, I'm nervous now. Tell you what, I knew Harmison was gonna enjoy this pitch. Good start! \\
\hline
\end{tabularx}
\end{table*}

\begin{table*}
\centering
\caption{Top anticipation Facebook posts}
\label{anticipationfb}
\begin{tabularx}{\textwidth}{|l|X|}
\hline
Score & Post \\
\hline
1.4652 & Please keep *PROPNAME*'s family in your prayers tonight and tomorrow. We have to be in court at 9am. Really pray for the oldest *PROPNAME* went to his dad's for his summer visit and had his mother, brother and sister stripped from him for six months. CPS is providing Christmas for the other two but not him.  I sure hope we succeed at stripping them of their   immunity   and make them pay for what they've done to this family.  \\
1.1556 & Dedicate yourself to the good you deserve and desire for yourself. Give yourself peace of mind. You deserve to be happy. You deserve delight. *PROPNAME* \\
1.0844 & Was on my first voice lesson today... feels good to start singing again... It was a long time ago... but now I'm back on track again ;) \\
1.0536 & long day of classes//library time... but.... *PROPNAME* is coming to see me tomorrow :)  So everything is good. \\
1.0536 & feels pretty ooky, but is going to try to power through and sing in everything tomorrow anyway. Good idea? Let's watch!  \\
1.033 & Seventh day of Christmas!!! I still hear the church bells ringing and Handel's Hallelujah Chorus!! Glory to God in the highest and on earth peace to all good men!!! \\
0.97647 & has contacts finally, its been a long time coming. its quite nice to see  \\
0.96156 & ohh, cleaning then reading for school... all the while just thinking about how much I can't wait to go home on Thursday!  Lots of time with my one and only, hunting, and all the fun and good people that go along with both!  Can Not Wait.  \\
0.96156 & Finally got a car! 1998 Nissan Altima GXE w// 98,700 miles on it in pretty good shape. Mechanics aside, glad I have a car w// a stereo again. Don't like that it's white, but I took what I could get! Might either mural-paint or shop-repaint it if it annoys me enough.  Also had a really great weekend (Thank you, guys. Xoxo!)--nice change from the last few! Glad as hell I'm not stranded anymore!!  \\
0.96156 & Hard drive problem?  Made funny noises twice and failed to boot.  Finally booted (obviously).  I am going to install a new one now.  Don't expect updates for a few minutes (hah!).  Of course the last time this happened, I installed a new drive, used the old one for backup, and the old continued working for years.  So this could all be a waste of time and money.  \\
\hline
\end{tabularx}
\end{table*}

\begin{table*}
\centering
\caption{Top joy tweets}
\label{joytweets}
\begin{tabularx}{\textwidth}{|l|X|}
\hline
Score & Tweet \\
\hline
7 & RE: High Road - Glad to help. U pray I pray We all pray. God provides. Our God is an awesome God! keep on keepin on \\
6 & i am becoming VERY NAUGHTY in good and bad way. Haha. I will just hugg and kiss everybody and dance dance dance dance : ... \\
6 & good luck good luck good luck! \\
6 & Peter Please Reply Me!! I Think That You Are Helping A Very Special Cause Good Luck And God Bless You! kiss! \\
5 & well hope its all you hope it is .....a lot of hope i realise but im quietly confident you will have a wonderful timexxx \\
5 & It won't be long until you find love, and by love, I mean TRUE love. You are such a beautiful, sweet, intelligent, \\
5 & love is energy yes - love is everything - the door, the chair, the tree, the house, the wind, God, to me love is everything \\
5 & hey! just wanted to say ive enjoyed praising god with you at church! You have an amazing gift! Im glad you are getting to share it \\
\hline
\end{tabularx}
\end{table*}

\begin{table*}
\centering
\caption{Top joy Facebook posts}
\label{joyfb}
\begin{tabularx}{\textwidth}{|l|X|}
\hline
Score & Post \\
\hline
1.2444 & Define me urban dictionary   The Most wonderful person in the World. Kind, Sweet, Loving, Caring, Gentle. Perfect in Every way. The one you love for all your life. Crazy hot girl. Beautiful, smart \& funny; *PROPNAME* posesses atributes absent in 99.9\% of women. Truly a lucky find. Plus she rocks. The most wonderful drug in the world, better know as lortab. taking the pill may cause a sense of euphoria, \& well being.   hahaha  \\
1.2319 & Let me know you more deeply and truely, oh Lord. There are none who know you completely, and so I pray that YOU show me. Your ways are perfect and I want my life to sing your praise. Let me walk with love, joy, peace, patience, kindness, goodness, gentleness, and self-control. Against such things there is no law.  \\
1.1556 & is thankful for everyone and everything. My life is beautiful and I am such a lucky girl. ? \& ?...Happy Thanksgiving! Love you all! \\
1.1484 & Happy Thanksgiving everybody! I'm thankful for friends to share Thanksgiving with:) Now time to make some yummy food to share later \\
1.1201 & is wishing a Happy Birthday to my bestie *PROPNAME*!   I love you honey, and I hope you have a wonderful 22nd Birthday!    \\
1.1201 & wishing my fiance a Happy, Happy Birthday!!  I love you darlin' and hope you have a wonderful day!  \\
1.033 & Seventh day of Christmas!!! I still hear the church bells ringing and Handel's Hallelujah Chorus!! Glory to God in the highest and on earth peace to all good men!!! \\
1.033 & Wish you all a VERY HAPPY 'Christmas'..enjoy with your family and friend..spread the smile.. Look in your socks in the morning and you ill find a gift - from Him.. Lets spread the Joy nd pray to let peace reign everywhere in the world.. \\
0.9768 & Please keep *PROPNAME*'s family in your prayers tonight and tomorrow. We have to be in court at 9am. Really pray for the oldest *PROPNAME* went to his dad's for his summer visit and had his mother, brother and sister stripped from him for six months. CPS is providing Christmas for the other two but not him.  I sure hope we succeed at stripping them of their   immunity   and make them pay for what they've done to this family.  \\
0.97647 & joy and freedom in the love of God! \\
\hline
\end{tabularx}
\end{table*}

\begin{table*}
\centering
\caption{Top joy headlines. The human-ranked headlines do not seem particularly
joyful; but nor does the 3rd lexically ranked headline, misled by
the ambiguity of 'baby' and 'found', but the first lexically ranked
headline is indisputably joyful.}
\label{joyheads}
\begin{tabularx}{\textwidth}{|l|l|X|}
\hline
Method & Score & Post \\
\hline
 & 93 & Goal delight for Sheva \\
Human & 93 & Lily Allen wins web music award  \\
 & 91 & China Successfully Launches Two Satellites \\
\hline
 & 3 & News Baby pandas! Baby pandas! Baby pandas! \\
Lexical & 3 & 'Jackass' star marries childhood friend The secrets people reveal \\
 & 3 & Outdated baby food found on shelves \\
\hline
\end{tabularx}
\end{table*}

\begin{table*}
\centering
\caption{Top surprise tweets. These suggest that 'hope' and 'luck'
are over-associated with surprise in the lexicon.}
\label{surprisetweets}
\begin{tabularx}{\textwidth}{|l|X|}
\hline
Score & Tweet \\
\hline
6 & good luck good luck good luck! \\
5 & Ok-I am surprisingly zero'd out.time to catch an early one;I have a feeling I'm going to lose this weekend as well.nite nite all!tweet good! \\
5 & Mosley can suck ma' balls. Good on FOTA for putting their money where their mouth is. I hope FIA caves and we get goodness. \\
4 & well hope its all you hope it is .....a lot of hope i realise but im quietly confident you will have a wonderful timexxx \\
4 & user10 Hiya Moira. Hope you had a lovely weekend. Did you have good weather down south? Twas OK up here. Hope youre well xx =] \\
4 & hope u dont mind but using your lovely sunset shot of serpentine as screen saver the colours are lovely u have talent xxx \\
4 & I NEED TO LOSE WEIGHT MAFUCKERRRRRRRRRRRRRRRR.I WANT TO LOSE WEIGHT.FUCK FOOD FUCK FOOD.FUCK ALL GOOD FUCK. \\
4 & good, good. nahh, im just worried about everyone else. sorry i ruined it for you. i hope this wont wreck our friendship \\
4 & what's good young lady?? Hope ur feeling great today...whatcha doing this weekend??? \\
4 & Drive a crappy car to deal so they know you have zero money... Then dress in clothes that are obviously too small. Good luck! \\
\hline
\end{tabularx}
\end{table*}

\begin{table*}
\centering
\caption{Top surprise Facebook posts. Again the association of the words
'death' and perhaps 'wishes' with surprise seems too strong; but the
7th ranked post shows that our normalization strategy is effective.}
\label{surprisefb}
\begin{tabularx}{\textwidth}{|l|X|}
\hline
Score & Post \\
\hline
0.82979 & finally feeling somewhat moved in. It might be a miracle folks! \\
0.77958 & Practice question for my EMT exam   its early afternoon and the alarm sounds You are called to a scene at a railroad one of the workers is trapped between 2 cars. He is alert and tells you that as soon as the cars are moved he will die. He asks that you leave him alone with his wife and to do nothing heroic. which emotional response to death is your patient exhibiting?   ...uhhhhh how intense is that?!  \\
0.72296 & The most beautiful experience we can have is the mysterious.  It is the fundamental emotion which stands at the cradle of true art and true science.  Whomever does not know it and can no longer wonder, no longer marvel, is as good as dead, and his eyes are dimmed. -Einstein  \\
0.719 & is wishing a Happy Birthday to my bestie *PROPNAME*!   I love you honey, and I hope you have a wonderful 22nd Birthday!    \\
0.69317 & thanks for the birthday wishes \\
0.69317 & Nordea DEATH. \\
0.69317 & EARTHQUAKE!!!! \\
0.69317 & My mouth always gets me in trouble... \\
0.69317 & feels like Death warmed over... \\
0.69317 & scored about 65k in bejeweled blitz in one second.  That was nice. \\
\hline
\end{tabularx}
\end{table*}

\begin{table*}
\centering
\caption{Top surprise headlines. Neither the human-ranked or lexically-ranked
headlines seem particularly surprising.}
\label{surpriseheads}
\begin{tabularx}{\textwidth}{|l|l|X|}
\hline
Method & Score & Post \\
\hline
 & 87 & Man rides stationary bike for 85 hours \\
Human & 79 & Archaeologists find remains of couple locked in a hug \\
 & 77 & Kathmandu: First snow in 63 years \\
\hline
 & 3 & PM: Havana deal a good experiment \\
Lexical & 2 & 4th person dies after receiving flu shot \\
 & 2 & Hungary police tackle violent protestors \\
\hline
\end{tabularx}
\end{table*}

\begin{table*}
\centering
\caption{Top anger tweets}
\label{angertweets}
\begin{tabularx}{\textwidth}{|l|X|}
\hline
Score & Tweet \\
\hline
7 & Is fed up. Bloody no job. Bloody no money. Bloody hospital. Bloody drugs. Bloody boredom. Bloody buggering bollocks. \\
5 & hell yeah that shit bomb i got everybody n the hood doin that shit \\
5 & u ain't never lie I seen 3 white men in the ceelo spot smfhhh!! Shit crazy they tryin to rape use for money land \& bitches!! Lmao \\
4 & damn how yall mad when i gave yall 2months to get yall shit together im backkkk lmfaoo my cocky shit yall like to hear \\
4 & Joe Jackson is a damn DISGRACE. This man is promoting a new record company and his son, Michael, hasn't been dead for 2 days. DAMN DISGRACE \\
4 & Yo Bobby Digital is stuck in my head Lol she be like "Nigga, You Aint Shit..Ya Daddy Aint Shit, Ya Money Aint Shit" Lol im bout to bump it \\
4 & Ok. 1 vote for LAC, 1 vote for Minny, and 1 vote for Bucks so far. C'mon guys, we need more NBA hell teams. \\
4 & I hate it when a girl runs from you screaming "RAPE!" when all I intended to do was eat her down to a screaming skeleton to gain her power. \\
\hline
\end{tabularx}
\end{table*}

\begin{table*}
\centering
\caption{Top anger Facebook posts. The 10th ranked post is an example
of mock anger, and the technique fails to detect the ironic language.}
\label{angerfb}
\begin{tabularx}{\textwidth}{|l|X|}
\hline
Score & Post \\
\hline
0.94206 & Recidivism (n.) - 1. Committing new offenses after being punished for a crime.  2.  Chronic repetition of criminal or other antisocial behavior. Replace   criminal   with   stupid  , and I think I've found my problem.  \\
0.91955 & s wireless internet keeps spazzing out every five minutes.  It's really annoying.  Actually, it's more than annoying, but I haven't gotten angry enough to start using foul language.  \\
0.77039 & Intelligence is a curse, life a senseless disease.  Gods grant me respite by death or lobotomy...  \\
0.73236 & Parent's Evening + Bomb Scare = Mucho Chaos. \\
0.71499 & is way too upset with the damn Dodgers to even really be upset. \\
0.71487 & Feeling DAMN GOOD headwise lately: forceful self-change of outlook, meds seem to be finally working, \& I'm startin to get shit done! 'Course, meds need tweaking--one of 'em (dunno which) is also tryin to make me a zombie--not the fun movie kind, either: I get sleepy around 4 or 5 PM now; too damn early even if I weren't a night owl! PSYCHED bout the forward progress, but this zombie thing's crampin' my nightlife, oy!  \\
0.7 & 8 pages in 2 hours...not bad. well, really bad actually! extremely bad what the hell am I saying   not bad   hahahaaaaaaaaaa  \\
0.69317 & and mosquito bites \\
0.69317 & Nordea DEATH. \\
0.69317 & down to 10 men and we're still kicking arse \\
\hline
\end{tabularx}
\end{table*}

\begin{table*}
\centering
\caption{Top anger headlines. Both sets are plausible, but again the
human-ranked list depends on implicit context rather than content.}
\label{angerheads}
\begin{tabularx}{\textwidth}{|l|l|X|}
\hline
Method & Score & Post \\
\hline
& 71 & Israeli woman's tirade spurs PM outrage \\
Human & 70  & Abbas condemns Israeli massacre in northern Gaza Strip \\
& 70  & Anger at release of two held over beheading plot \\
\hline
& 3 &  Poison Pill to Swallow: Hawks Hurting After Loss to Vikes \\
 Lexical & 3 &  Brazil Air Force Cites Faults and Confusion in Fatal Crash \\
& 3 &  Iraqi suicide attack kills two US troops as militants fight purge \\
\hline
\end{tabularx}
\end{table*}

\begin{table*}
\centering
\caption{Top disgust tweets. Note the overlap with high-ranked anger
tweets, reflecting that some emotions are not well separated by
the empirical approach to constructing the NRC lexicon.}
\label{disgusttweets}
\begin{tabularx}{\textwidth}{|l|X|}
\hline
Score & Tweet \\
\hline
6 & Is fed up. Bloody no job. Bloody no money. Bloody hospital. Bloody drugs. Bloody boredom. Bloody buggering bollocks. \\
5 & i think its because im sick. Doc got me blowin a balloon with my nose and shit... shit aint workin. This shit is madd annoying... \\
4 & Hell yea " So you callin me a liar!?" Shit hell yea! \\
4 & hell yeah that shit bomb i got everybody n the hood doin that shit \\
4 & i never sleep too.. its shit. yeah i been ill since wednesday night, been well bad, i been semi diagnosed with swine flu \\
4 & damn how yall mad when i gave yall 2months to get yall shit together im backkkk lmfaoo my cocky shit yall like to hear \\
4 & Joe Jackson is a damn DISGRACE. This man is promoting a new record company and his son, Michael, hasn't been dead for 2 days. DAMN DISGRACE \\
4 & ysl is the shit vavy @user3985 nigga i found all the hard yeve saint laurent shit the got shit for the low lol belts only 275 cop out boy! \\
4 & I hate it when a girl runs from you screaming "RAPE!" when all I intended to do was eat her down to a screaming skeleton to gain her power. \\
\hline
\end{tabularx}
\end{table*}

\begin{table*}
\centering
\caption{Top disgust Facebook posts. The 9th ranked post expresses many
emotions, but the disgust score comes from the use of `zombie'
and `shit'.}
\label{disgustfb}
\begin{tabularx}{\textwidth}{|l|X|}
\hline
Score & Post \\
\hline
0.79003 & worst night ever! then I get this message?   hey so i just moved up here and ill be honest, living with someone but not getting needs met sexually and hoping to find someone in same position or at least who understands and can be discrete, if you may be interested let me know ,if not i am very sorry to have bothered you   I could be a god damn serial killer who gets off setting people on fire urg I fucking hate people!  \\
0.77338 & Recidivism (n.) - 1. Committing new offenses after being punished for a crime.  2.  Chronic repetition of criminal or other antisocial behavior. Replace   criminal   with   stupid  , and I think I've found my problem.  \\
0.77041 & Heh...:  God I wish that I could hide away//And find a wall to bang my brains//I'm living in a fantasy,//a nightmare dream...reality//People ride about all day//In metal boxes made away//I wish that they would drop the bomb//And kill these cunts//that don't belong! I hate people!//I hate the human race//I hate people!//I hate your ugly face//I hate people!//I hate your fucking mess//I hate people!//They hate me!  -Anti-Nowhere League  \\
0.77039 & Intelligence is a curse, life a senseless disease.  Gods grant me respite by death or lobotomy...  \\
0.76996 & Well, lived through the Carbon Monoxide shit, though it made me sick today--luckily not sick enough to go to the damn hospital I guess; just sick enough to be puking and bedridden today. Good thing I had the day off anyway for MLK day (but my dad didn't tell me that, so I woke up ill and panicky this afternoon--nice, lol). Furnace was fixed today, though, so no more CM troubles, I suppose!  \\
0.7528 & I have a ship and a fighting crew and a girl with lips like wine, and that's all I ever asked.  Lick your wounds, bullies, and break out a cask of ale.  You're going to work ship as she never was worked before.  Dance and sing while you buckle to it, damn you! To the devil with empty seas! We're bound for waters where the seaports are fat, and the merchant ships are crammed with plunder!    \\
0.73236 & is tired of being sick. Death to all swine. \\
0.71499 & got sick of the crap £5 haircuts so upgraded to the £22 haircut today.still crap. \\
0.71487 & Feeling DAMN GOOD headwise lately: forceful self-change of outlook, meds seem to be finally working, \& I'm startin to get shit done! 'Course, meds need tweaking--one of 'em (dunno which) is also tryin to make me a zombie--not the fun movie kind, either: I get sleepy around 4 or 5 PM now; too damn early even if I weren't a night owl! PSYCHED bout the forward progress, but this zombie thing's crampin' my nightlife, oy!  \\
\hline
\end{tabularx}
\end{table*}

\begin{table*}
\centering
\caption{Top disgust headlines. The human-ranked list depends on knowing
that teachers, especially, are not supposed to assault others sexually,
and that child porn is especially disgusting.}
\label{disgustheads}
\begin{tabularx}{\textwidth}{|l|l|X|}
\hline
Method & Score & Post \\
\hline
 & 87 & Teacher charged with sex assault \\
Human & 74 & Abbas condemns Israeli massacre in northern Gaza Strip \\
 & 73 & Investigation of child porn site hits 77 nations \\
\hline
 & 2 &  Teen repellent inventor wins infamous prize  \\
Lexical & 2 &  Israeli woman's tirade spurs PM outrage \\
 & 2 &  Observatory: poison begets poison \\
\hline
\end{tabularx}
\end{table*}

\begin{table*}
\centering
\caption{Top fear tweets. The lexicon struggles with this emotion because
tweets tend to contain hyperbole: `die', `dying', and `drowning' are
not the fear-laden words in this context that the lexicon assumes they
are. Notice how much social context would be required 
to understand that these are not fear words in these tweets.}
\label{feartweets}
\begin{tabularx}{\textwidth}{|l|X|}
\hline
Score & Tweet \\
\hline
7 & Is fed up. Bloody no job. Bloody no money. Bloody hospital. Bloody drugs. Bloody boredom. Bloody buggering bollocks. \\
6 & RE: High Road - Glad to help. U pray I pray We all pray. God provides. Our God is an awesome God! keep on keepin on \\
5 & It's Godwin's Law.  It's also insulting, considering the last 8 years of Bush rule, I doubt these fuckers were screaming foul. \\
4 & agree  problematic as the "war on terror" has been labeled by some as the endless war \\
4 & This lil girl wants 2 watch spongebob squarepants bitch I'm 20 yrs old I'm grown either we watch Blues Clues or we won't be watchin tv @ all \\
4 & its going Great!! Keeps me Honest!! Its fun to look for the best bang for my buck!! Are u gonna watch BB11 tonight? \\
4 & I hate it when a girl runs from you screaming "RAPE!" when all I intended to do was eat her down to a screaming skeleton to gain her power. \\
4 & Chicken kebab from local chipper is to die for. No doubt dying is what I will be in the morning. \\
4 & One thing I don't mind drowning in is confidence. Hell, I live in a confidence desert. - -URL- \\
4 & I HATE awkward moments in movies. They make me switch tabs and do something else to avoid the imminent embarrassment. Why do I even bother? \\
\hline
\end{tabularx}
\end{table*}

\begin{table*}
\centering
\caption{Top fear Facebook posts. Again, the lexicon labels many
words as expressions of fear, when they are often used hyperbolically.
The labelling of `pray' as a fear word is understandable but, as
the 9th ranked post shows, is inappropriate.}
\label{fearfb}
\begin{tabularx}{\textwidth}{|l|X|}
\hline
Score & Post \\
\hline
1.2175 &  Heh...:  God I wish that I could hide away//And find a wall to bang my brains//I'm living in a fantasy,//a nightmare dream...reality//People ride about all day//In metal boxes made away//I wish that they would drop the bomb//And kill these cunts//that don't belong! I hate people!//I hate the human race//I hate people!//I hate your ugly face//I hate people!//I hate your fucking mess//I hate people!//They hate me!  -Anti-Nowhere League  \\
1.0062 & (continued from above) And rot inside a corpse's shell The foulest stench is in the air The funk of forty thousand years And grizzly ghouls from every tomb Are closing in to seal your doom And though you fight to stay alive Your body starts to shiver For no mere mortal can resist The evil of the thriller (Into maniacal laugh, in deep echo)  \\
0.96299 & Intelligence is a curse, life a senseless disease.  Gods grant me respite by death or lobotomy...  \\
0.9037 & Many of us spend our whole lives running from feeling with the mistaken belief that you cannot bear the pain. But you have already borne the pain. What you have not done is feel all you are beyond the pain.     -*PROPNAME*  \\
0.88545 & I'm ecstatic! I just spoke briefly with *PROPNAME*'s lawyer. Based on some   case files   I emailed her, we have enough evidence to prove CPS's removal of the kids was illegal and grounds for an emergency hearing. We believe we have enough evidence to cost both CPS and Humble PD their   qualified immunity  , allowing us to file a civil suit against them and the apartments. ANYONE KNOW A GOOD CIVIL ATTORNEY???  \\
0.88246 & Put this as your status if you or somebody you know has suffered the LOSS of a Baby ? The majority won't put it on because unlike cancer. Baby Loss is a taboo... Break the Silence ? This is in memory of all the Angel Babies...gone too soon but NEVER FORGOTTEN ABOUT!!? Show your support and let these women know they don't have to grieve alone. \\
0.86085 & wonders, with all this righteous indignation over a terrorist assassination program and change in tactics from pursuing the Taliban to protecting the population, when did this stop being a war?  \\
0.8594 & I want your love and  I want your revenge     You and me could write a bad romance       I want your love and  All your lovers' revenge       You and me could write a bad romance    \\
0.85341 & Last night I went to a conference where the police were asking us to pray for them. Lets pray for the police and our communities. God moves when his people pray. \\
0.79003 & worst night ever! then I get this message?   hey so i just moved up here and ill be honest, living with someone but not getting needs met sexually and hoping to find someone in same position or at least who understands and can be discrete, if you may be interested let me know ,if not i am very sorry to have bothered you   I could be a god damn serial killer who gets off setting people on fire urg I fucking hate people!  \\
\hline
\end{tabularx}
\end{table*}

\begin{table*}
\centering
\caption{Top fear headlines. The human-ranked list is better, but again
because of knowledge about, for example, letter bombs as a tool of
terrorism.}
\label{fearheads}
\begin{tabularx}{\textwidth}{|l|l|X|}
\hline
Method & Score & Post \\
\hline
 & 95 & New Iraq terror tape calls for abducting foreigners \\
Human & 94 & Bombers kill shoppers \\
 & 92 & UK workers on alert for letter bombs \\
\hline
 & 5 & Freed kidnap suspect: my terror at police raid \\
Lexical & 4 & Brazil Air Force Cites Faults and Confusion in Fatal Crash \\
 & 4 & Iraqi suicide attack kills two US troops as militants fight purge \\
\hline
\end{tabularx}
\end{table*}

\begin{table*}
\centering
\caption{Top sadness tweets. The associations of `black', `blues', and 
`nothing' with sadness explain a number of these high-ranked tweets,
but they are common words for which their other meanings are
not sad.}
\label{sadnesstweets}
\begin{tabularx}{\textwidth}{|l|X|}
\hline
Score & Tweet \\
\hline
8 &  Is fed up. Bloody no job. Bloody no money. Bloody hospital. Bloody drugs. Bloody boredom. Bloody buggering bollocks. \\
5 &  Oh please. Dark romance. I love me some dark romance. Just the kind of news to make my bad day better :-) \\
4 &  You were everything thats bad for me, make no apology. Im crushed black and blue. But you know id do it all again for you. \\
4 &  Ok. 1 vote for LAC, 1 vote for Minny, and 1 vote for Bucks so far. C'mon guys, we need more NBA hell teams. \\
4 &  user4461 just in case u were worried I didn't die of heat stroke or get kidnapped! Lol \\
4 &  Nothing goes on the show without testing by the Black and Blues RT user4734 user4739 do the black and blues even test the motivator? \\
4 &  Is it better to be sick on a day off and lose the day but not have to do anything, or be sick on a show day and lose the day off? Hmmm \\
4 &  *rabid foam kill kill kill* :-) \\
4 &  The cake is a lie. -NL-The cake is a lie. -NL-The cake is a lie. -NL-The cake is a lie. \\
4 &  user6636 good unfair and bad unfair for me. At Crawford. Much worse hours. \\
\hline
\end{tabularx}
\end{table*}

\begin{table*}
\centering
\caption{Top sadness Facebook posts. For longer documents, the sadness
lexicon seems to perform better.}
\label{sadnessfb}
\begin{tabularx}{\textwidth}{|l|X|}
\hline
Score & Post \\
\hline
0.9037 & Attention everyone: If you have the misfortune of reading this, then I regret to inform you that I've been seriously injured in a car accident. Okay, maybe not, but I did just lose the game....  \\
0.86247 & (continued from above) And rot inside a corpse's shell The foulest stench is in the air The funk of forty thousand years And grizzly ghouls from every tomb Are closing in to seal your doom And though you fight to stay alive Your body starts to shiver For no mere mortal can resist The evil of the thriller (Into maniacal laugh, in deep echo)  \\
0.84339 & Recidivism (n.) - 1. Committing new offenses after being punished for a crime.  2.  Chronic repetition of criminal or other antisocial behavior. Replace   criminal   with   stupid  , and I think I've found my problem.  \\
0.82789 & Sick of this nihilistic depression shit--fine, life may be meaningless, I may be right about all the other disturbingly bleak shit I've been thinking...but I'm not a self-killer so I'm stuck here--might as well pull it together and live this crazy life I've got anyway, eh?  Some change is in the works (just in time for the New Year, ironically).  \\
0.77041 & Heh...:  God I wish that I could hide away//And find a wall to bang my brains//I'm living in a fantasy,//a nightmare dream...reality//People ride about all day//In metal boxes made away//I wish that they would drop the bomb//And kill these cunts//that don't belong! I hate people!//I hate the human race//I hate people!//I hate your ugly face//I hate people!//I hate your fucking mess//I hate people!//They hate me!  -Anti-Nowhere League  \\
0.77039 & Intelligence is a curse, life a senseless disease.  Gods grant me respite by death or lobotomy...  \\
0.75543 & Hmm, that wasn't too bad. Had me worried for a minute.  I've certainly seen worse endings. Like freakin' Fallen! Seriously, what the crap was that?! Anyways, not bad. \\
0.74509 & Gotta love 3 mobile. I asked to cancel my contract and was asked   could we convince you to stay   I replied with   Due to the terrible customer service, I would rather die a slow and horribly painful death than stay   to which I got   do you have any friends or relatives you would like to refer so they can get a good deal  . I want that CSR to work in my company retention department!!  \\
0.74453 & will gladly boast in his weaknesses. When I am weak, then He is strong. Your might, oh Lord is endless. Please cover over me. Please cover over your beloved..Lest we  all fall in overwhelming shame under the weight of our imperfections.  \\
0.73236 & ....lost my bromley id. already. curse my absent mind. \\
\hline
\end{tabularx}
\end{table*}

\begin{table*}
\centering
\caption{Top sadness headlines. The human-ranked list is better;
all three of the lexically-ranked list fail to grasp implicit content.}
\label{sadnessheads}
\begin{tabularx}{\textwidth}{|l|l|X|}
\hline
Method & Score & Post \\
\hline
 & 96 & Iraqi death toll exceeded 34,000 in 2006, United Nations says \\
Human & 95 & Iraq car bombings kill 22 People, wound more than 60 \\
 & 94 & Participant in water-drinking contest dies \\
\hline
 & 3 & Poison Pill to Swallow: Hawks Hurting After Loss to Vikes \\
Lexical  & 2 & Injured Marathon Winner Leaves Hospital \\
 & 2 & Beating poverty in a small way \\
\hline
\end{tabularx}
\end{table*}

\subsection{Wishes}

Goldberg \emph{et al.} investigated whether wishes or desires could
be detected lexically. Their goal was to go beyond the limitations of
sentiment analysis (how does a consumer feel now?) to find 
objects that a consumer might feel positively towards 
even if they were not actually present.
They developed models for the language of wishes based both on words 
themselves, and a set of templates
(e.g. `I just want $\ldots$').

We created a wish lexicon from their corpus in which each document
is labelled as embodying a wish or not. We extracted all parts of
speech except nouns, and selected words based on their significance
as predictive attributes for the wish/non-wish class label.
This produced a set of 28 words that are associated with the
wish/non-wish distinction. The lexicon is shown in Table~\ref{wishlexicon}.

\begin{table*}
\centering
\caption{Wish lexicon}
\label{wishlexicon}
\begin{tabularx}{\textwidth}{|X|}
\hline
 \\
again, 
buy, 
could, 
find, 
get, 
getting, 
going, 
hope, 
hoped, 
hopefully, 
hoping, 
lack, 
like, 
liked, 
looking, 
love, 
need, 
now, 
own, 
please, 
pleases, 
should, 
take, 
use, 
want, 
wanted, 
wish, 
would \\
 \\
\hline
\end{tabularx}
\end{table*}

We then repeated the scoring process used for measuring
emotions with this new lexicon,
getting word counts for each document, normalizing for
document length as before, and summing the adjusted counts.
Table~\ref{wishestweets} shows the top-ranked tweets and
Table~\ref{wishesfb} the top-ranked Facebook posts.

\begin{table*}
\centering
\caption{Top tweets embodying wishes. This lexicon does seem to perform
reasonably at detecting wishes, even ones that are quite implicit.}
\label{wishestweets}
\begin{tabularx}{\textwidth}{|l|X|}
\hline
Score & Tweet \\
\hline
5 & What's your favorite love song? I need to find good ones to download! I need 2 feel the love! \\
5 & religion promotes alot of hate and does more harm then good sometimes! Do want you want, with who you want, how you want! \\
5 & omg love penry-jones!!  my fav austen adaptation - have watched over 50 times.  love love love.  so much love. \\
5 & I know u want ME!.. i KNOW u want meee u know i want that XD! ...  user11735 man are u crazy or what!?.. I love to have sex.. i dont want SIDA! \\
5 & What do you like better tweed or spaz I'm thinking I like tweed better I'm the only one in my circle with the pre need dif view \\
5 & I want a free iPhone so bad I can taste it. ...it tastes less like an Apple and more like plastic and desperation.\#HGBW \#FreeiPhone3GS \\
4 & May love surround you and those you love! Be the love that you are and all will unfold better than you can imgaine! \\
4 & You guys were amazing! I love the new songs! I love you more and more after each show! :-) \\
4 & Hit the road, Wes, and dontcha come back NO MORE NO MORE NO MORE NO MORE! \#Bachelorette \\
4 & *hugs* do hope you fall asleep quickly - sounds like you need more rest.  Hope you feel better soon. : ) \\
4 & I, too, love the new vlog :) Keep 'em comin'!! You are both so adorable and I want more too! :D Love ya!!! \\
4 & Less like a date, more like spending time with an old friend. \\
4 & Oh I don't think it's gross, it's more like I want to high-five him.  'Sides, ain't like it was uncommon at the time. \\
4 & so sorry but i love you... (love you more, more) \\
4 & user7808 i'd rather watch will and grace. HA! i want primo, i want DVDs, i want everything. \\
4 & Shaft- do you want the ld or do you want to be held? Random woman- i want the ld then i want to be held. \\
\hline
\end{tabularx}
\end{table*}

\begin{table*}
\centering
\caption{Top Facebook posts embodying wishes}
\label{wishesfb}
\begin{tabularx}{\textwidth}{|l|X|}
\hline
Score & Post \\
\hline
1.3207 & Gotta love 3 mobile. I asked to cancel my contract and was asked   could we convince you to stay   I replied with   Due to the terrible customer service, I would rather die a slow and horribly painful death than stay   to which I got   do you have any friends or relatives you would like to refer so they can get a good deal  . I want that CSR to work in my company retention department!!  \\
1.2643 & is making an attempt to brief a few cases while waiting for *PROPNAME* to get off work so she can talk to him... I think I'm actually going to like law school, but I wish my love could be here with me! I miss him :(  \\
1.1769 & Fucking CARBON MONOXIDE alarm's going off repeatedly again tonight! Opened the windows \& garage doors; I have nowhere else I can go \& I need sleep, so I'm gonna hafta just hope the ventilation is enough cos I don't know what else to do. Hope I don't end up in the hospital over this shit--I'm sick of the damn hospital! (Well...bein' dead would [debatably] suck too. But I wouldn't be self-aware to know about it, lol.)  \\
1.1769 & Really?  Two days in a row?  I would like to take back my comment about de ja vu - I didn't mean I wanted it to end like that (again). \\
1.1218 & How messy my room is again.. i love searching for random thngs.. and how nice bed is looking right now! Ninite all, hope you had great dreams and better sleeps!  \\
1.1218 & worst night ever! then I get this message?   hey so i just moved up here and ill be honest, living with someone but not getting needs met sexually and hoping to find someone in same position or at least who understands and can be discrete, if you may be interested let me know ,if not i am very sorry to have bothered you   I could be a god damn serial killer who gets off setting people on fire urg I fucking hate people!  \\
1.0894 & Dear God, I know you've got a lot going on with Jesus' birthday coming up and all, but I need you to take this Civil Procedure exam for me in the morning.  I know you take all of my exams for me, but I really need ya in an extra special sort of way for this one. Oh and  one more thing, please hold off the snow until I can make it home....  \\
1.0894 & If I should stay, I would only be in your way. So I'll go, but I know I'll think of you ev'ry step of the way...I hope life treats you kind. And I hope you have all you dreamed of. And I wish you joy and happiness but above this I wish you love....And I   \\
1.0536 & thinks that no one should be afraid to go to the doctor because they can't afford it, no one should go broke because they get sick, and no one should die because they could not afford care. If you agree, please post this as your status for the rest of the day.  \\
1.0536 & wishes life weren't so gray-scale... There are things I fear that excite me, things I want that I'm not sure I want, things I'm doing that I wish I were not, things I should be doing that I dread, and things that I want to put off but can't wait to get over with.  WTH life, how about giving me 2+2 instead of [x\^3-x\^2]//[(x\^2+9)\^3dx]?  \\
1.0536 & Well, see what you want to see, you should see it all. Well, take what you want from me, you deserve it all. Nine times out of ten our hearts just get dissolved. Well, I want a better place or just a better way to fall.  \\
\hline
\end{tabularx}
\end{table*}

\subsection{Preferences}

The same process was repeated with a lexicon manually constructed
to capture aspects of desire and preference.
The lexicon of desires and preferences is shown in Table~\ref{desirelexicon}.
The difference between wishes and preferences is that preferences
usually refer to either the current, or imminent, situation, while
wishes usually indicate a longer-range desire or preference.
A robot might process a desire or preference immediately, while storing
a wish as part of its model of the corresponding human.

\begin{table*}
\centering
\caption{Preference/Desire lexicon}
\label{desirelexicon}
\begin{tabularx}{\textwidth}{|X|}
\hline
 \\
like, 
dislike, 
love, 
want, 
need, 
prefer, 
rather, 
better, 
worse, 
more, 
less, 
favourite, 
favorite \\
 \\
\hline
\end{tabularx}
\end{table*}

Table~\ref{desirestweets} shows the top-ranked tweets and
Table~\ref{desiresfb} the top-ranked Facebook posts for this model.

\begin{table*}
\centering
\caption{Top tweets expressing preferences}
\label{desirestweets}
\begin{tabularx}{\textwidth}{|l|X|}
\hline
Score & Tweet \\
\hline
5 & What's your favorite love song? I need to find good ones to download! I need 2 feel the love! \\
5 & religion promotes alot of hate and does more harm then good sometimes! Do want you want, with who you want, how you want! \\
5 & omg love penry-jones!!  my fav austen adaptation - have watched over 50 times.  love love love.  so much love. \\
5 & I know u want ME!.. i KNOW u want meee u know i want that XD! ...  @user11735 man are u crazy or what!?.. I love to have sex.. i dont want SIDA! \\
5 & What do you like better tweed or spaz I'm thinking I like tweed better I'm the only one in my circle with the pre need dif view \\
5 & I want a free iPhone so bad I can taste it. ...it tastes less like an Apple and more like plastic and desperation.\#HGBW \#FreeiPhone3GS \\
4 & May love surround you and those you love! Be the love that you are and all will unfold better than you can imgaine! \\
4 & You guys were amazing! I love the new songs! I love you more and more after each show! :-) \\
4 & Hit the road, Wes, and dontcha come back NO MORE NO MORE NO MORE NO MORE! \#Bachelorette \\
4 & *hugs* do hope you fall asleep quickly - sounds like you need more rest.  Hope you feel better soon. : ) \\
\hline
\end{tabularx}
\end{table*}

\begin{table*}
\centering
\caption{Top Facebook posts expressing preferences}
\label{desiresfb}
\begin{tabularx}{\textwidth}{|l|X|}
\hline
Score & Post \\
\hline
0.97647 & Gotta love 3 mobile. I asked to cancel my contract and was asked   could we convince you to stay   I replied with   Due to the terrible customer service, I would rather die a slow and horribly painful death than stay   to which I got   do you have any friends or relatives you would like to refer so they can get a good deal  . I want that CSR to work in my company retention department!!  \\
0.97647 & I need to tell people I'm single more often. Two girls talk to me in less than 24 hours. That's like twice as many in the last week. Wow. I feel special. That's why I ride the short bus. \\
0.83631 & My love, I forgive you; you never planned to die - and love, I'll place two pennies over your eyes... And I will love you, after the war. I love you for always, forever more. I will love you, after the war - forever, for always, and more...    \\
0.82979 & is going to need a lot more upper body strength if I ever want to be a pole dancer \\
0.82979 & Going off to kick my own ass at the gym once more.... For better or worse, this type of masochism is highly addictive  \\
0.82979 & The day you wake up and realize you've just slept twelve hours and still want to sleep some more is the day you need to get back on a regular sleeping schedule. \\
0.82979 & is being conformed to the image of Christ. Do you want to be like Jesus? Then you must learn something...you must learn how to love someone who does not meet the conditions. It's not easy, but it's not impossible.  \\
0.82979 & Let me know you more deeply and truely, oh Lord. There are none who know you completely, and so I pray that YOU show me. Your ways are perfect and I want my life to sing your praise. Let me walk with love, joy, peace, patience, kindness, goodness, gentleness, and self-control. Against such things there is no law.  \\
0.82979 & ...my doctor says I have   a cute virus  or something like that Bahahahaha No H1N1 here yet! Bath, lots of liquids and I might even take a nap. Need to get better for date night with *PROPNAME* tomorrow!  \\
0.82979 & Dear Online Dictionary Makers, Let people search the text of the entries, not just the entry words.  Make your dictionaries more like a human brain.  Let me search   make things the same   and find words that include those words in their definitions.  That would rock. Love, *PROPNAME*  \\
\hline
\end{tabularx}
\end{table*}

\subsection{Personality}

The essays and Facebook posts corpora are
labelled by the Big Five personality properties
of each author, so we can build and assess predictive models for 
each of the Big Five dimensions.
We build and test models for the essays, Youtube, and Facebook corpora.
For the first two corpora, this represents a few hundred words,
and so stands in for what a robot could collect over the course
of several interactions with a human. The Facebook corpus, on the
other hand, stands in for a single, short interaction for which it
might be expected to be quite difficult to infer personality traits.

Our process is as follows. For each extreme of a personality
dimension (e.g. for Extroversion: extrovert, and introvert) we select 
as training data documents authored by individuals with that trait.

For example, to predict extroversion versus introversion,
we select 100 randomly chosen documents authored by
extroverts and 100 authored by introverts. 
For each set, we construct a document-word matrix based on the
most-frequent 1000 words, and normalize based on the log + digamma
process described previously.
This gives us a $100 \times 1000$ matrix, $A$,
whose rows represent documents, columns represent words, and entries
represent normalized word pseudo-rates. 
We normalize the columns of this matrix to
z-scores, recording the mean and standard deviation of each column
so that we can normalize test records to match.

We compute the singular value decomposition
of $A$ as
\[
A \;\approx\; U S V'
\]
where $U$ is an $n \times k$ matrix representing the documents in
$k$-dimensional space, $S$ is a diagonal matrix with non-increasing
entries representing the amount of variation in each dimension, and
$V$ is a $1000 \times k$ matrix representing the variation among the
words.

Given a test record, expressed as word pseudo-rates, we apply the means
and standard deviations used for the training data to map it to a
scale that matches the training data. Then we recompute the SVD
with the test record added to the end of $A$, compute a new
$V$ matrix, $\hat{V}$, which is still $v \times k$, and compute
the Frobenius norm difference between $V$ and $\hat{V}$.
(In fact, we do not need to recompute the SVD; see the appendix.)

If the test records `fits' with the model built from the original
training set, then the difference in the Frobenius norm will tend to
be larger than if it does not.
This can be seen by considering the properties of the SVD. One
interpretation of an SVD is that, if the rows of $U$ and the rows of
$V$ are plotted in the same $k$-dimensional space, the distance
between a point representing a row of $U$ and a point representing
a row of $V$ corresponds to the affinity between that object and
that attribute. Adding a new point corresponding to an added row of
$U$ has a different effect, depending on how similar it is to the
existing rows.
If it is like them, then it exerts strong pulls on all of the
attribute points because it, like the rest of the object points,
is close to them. If, however, it is not like the existing objects,
then it is plotted far from them, and it is therefore far from
the attribute points as well. Hence, it exerts only a weak pull on
them; which is reflected in a small change in norm.

We have no scale that can be used to determine when a norm change
is large or small. We use both of the models 
from the extremes of each personality dimension, 
the models for extroverts and introverts. Given a test point, 
we predict it to belong to the class (extreme) that claims it most strongly.

Table~\ref{predictbig5} shows the performance of predictors built
in this way for all five personality dimensions, and for the essays,
Youtube, and Facebook datasets. There are no directly comparable
results using other techniques, since these tend also to use other
syntactic markers, but most achieve F-scores in the 60\% range,
while ours are typically higher, sometime much higher.

There are interesting differences across the dataset. For the essays
dataset, prediction accuracy for all five dimensions is comparable;
for the Youtube dataset, prediction of the conscientious dimension
is noticeably higher than the others, but all are higher than for
the essays. For the Facebook dataset, prediction accuracy
for openness is higher than for the other dimensions, but all are lower
than for the other two datasets. This is not surprising given that
Facebook posts are relatively short; indeed, what is surprising is
that aspects of personality can be predicted from such short texts.
We doubt the many humans would feel confident about doing so.

\begin{table*}[]
\def\arraystretch{1.1}%
\centering
\caption{Performance predicting Big Five personality dimensions}
\label{predictbig5}
\begin{tabulary}{\textwidth}{|C|C|C|C|C|C|C|}
\hline
Dataset/Personality dimension & Score & Openness & Conscientiousness & Extraversion & Agreeableness & Neuroticism \\ \hline
\multirow{2}{*}{Essays}       & Av. & 0.71     & 0.69              & 0.71         & 0.70          & 0.70        \\ \cline{2-7} 
                              & F1  & 0.73     & 0.72              & 0.72         & 0.71          & 0.72        \\ \hline
\multirow{2}{*}{Youtube}      & Av. & 0.79     & 0.92              & 0.85         & 0.79          & 0.80        \\ \cline{2-7} 
                              & F1  & 0.81     & 0.92              & 0.85         & 0.80          & 0.80        \\ \hline
\multirow{2}{*}{Facebook}     & Av. & 0.79     & 0.65              & 0.67         & 0.63          & 0.73        \\ \cline{2-7} 
                              & F1  & 0.82     & 0.72              & 0.74         & 0.70          & 0.76        \\ \hline
\end{tabulary}
\end{table*}

\subsection{Semantic differential}

Osgood postulated that there was a natural three-dimensional structure
underlying human value judgements, in which all words could be considered
to lie at coordinate positions with respect to three independent
conceptual axes, which he named the semantic differential \cite{osgood}. 
He suggested that words varied along the dimensions
of good--bad, active--passive, and strong--weak (sometimes
called Evaluation, Activity, and Potency). This construct was
validated extensively across cultures, beginning with adjectives
but then extended to a range of other words \cite{osgood:cultural}.
Osgood's view of this three-dimensional structure was attitudinal
in flavor: ``The way the lexicon carves up the world" (p171).
The semantic differential approach gave rise to a line of research that 
still continues in marketing, and led indirectly to the popularity of 
Likert scales for eliciting opinions.
Russell and Mehrabian \cite{russell:mehrabian} characterize essentially
the same three dimensions as underlying the full spectrum of emotions,
but this is not surprising since Osgood's claim was that these three
dimensions are ``a universal framework underlying certain affective or 
connotative aspects of language" (Ibid).

This semantic differential is one conceivable way to begin to
model attitudes (and perhaps eventually values).
In our context, the semantic differential scores of an utterance
can be considered as capturing the attitudes of the speaker to 
the general immediate context.
Both tweets and Facebook posts can be considered as declarative
statements of the worldview of the author at the moment that they
are uttered. Some have an obvious target in mind; others are simply
general comments on life as perceived by the author.
In the context of Facebook this is explicit in the naming of the
post as a `status update'.

It is possible to imagine a robot inferring more sophisticated
attitudes by computing the current object of attention of a human in its
vicinity (for example, using gaze), 
and associating attitudinal markers in speech with
this object. This is still an area in its infancy.

We take the same corpora previously used to infer emotions and
compute the three intensities of each document on the three Osgood
dimensions. Heise \cite{heise} computed loadings on each of the 
three dimensions for the most common 1000 words in English. 
We construct document-word
matrices from the documents in each corpora as before, and
weight each word by the corresponding loading for each of the three
dimensions. We then score each document by adding up the weights of
each word it contains. The loadings have values between $-1$ and $+1$,
so the most intense document in each dimension will have a large
positive score, and the least intense a large negative score.

These scores reflect the attitude of the author to the external
world at the instant it was written. So, for example,
`good' reflects the author's view of his/her context, 
as opposed to positive mood, which reflects the author's internal state.
As before, we list the top ranked at each extreme to illustrate.
These are shown in 
Table~\ref{goodtweets} (top good tweets),
Table~\ref{goodfb} (top good Facebook posts),
Table~\ref{badtweets} (top bad tweets),
Table~\ref{badfb} (top bad Facebook posts),
Table~\ref{activetweets} (top active tweets),
Table~\ref{activefb} (top active Facebook posts),
Table~\ref{passivetweets} (top passive tweets),
Table~\ref{passivefb} (top passive Facebook posts),
Table~\ref{strongtweets} (top strong tweets),
Table~\ref{strongfb} (top strong Facebook posts),
Table~\ref{weaktweets} (top weak tweets), and
Table~\ref{weakfb} (top weak Facebook posts).

\begin{table*}
\centering
\caption{Top good attitude tweets. Because the lexicon is large, and
each word has a weighting between $-1$ and $+1$, it is harder to
see why each document scores as it does. However, only the 3rd ranked
tweet is hard to understand.}
\label{goodtweets}
\begin{tabularx}{\textwidth}{|l|X|}
\hline
Score & Post \\
\hline
7.00 & Good Morning!! Come here and I will give you something to do! Really fast!! haha. I'm here until 2 pm sharp!! :D \\
6.52 & That's good because we only wear all black to express our inner angst. \\
6.43 & Much more history here. I'm staying at the pub Admiral Nelson was at when he was called up to fight in what became the battle of Trafalgar. \\
6.31 & dragon ball z was that much more DOPE... back in the daay. narutos not really my thing... ironing ? good boy : ) \\
5.74 & Twitter-ville, you got a good idea for a title for my Special/DVD/CD? I got nothing I really like and need one tomorrow. \\
5.73 & Car nice and clean for later on! Got my tent fot T! HAd it up and down, packed away in 10 mins! :) How is every1? x \\
5.65 & yeah poor little baby rosco will have to be in the way back all by himself u need to light a fire under my butt about my blog lol \\
5.57 & I will be missing the Bayview Social as I will be volunteering at the Wheelchair stand at Festa Italiana! Got to give back! Enjoy! \\
5.52 & thank u so much user3965 for always being such a good friend \& tweeting about the show i'm producing user4611 w user4612 Love u Kim! \\
5.40 & Only to u... I hav been chatting over there no problem, I must really do some study cause hav not done any since b4 Melbourne \\
\hline
\end{tabularx}
\end{table*}

\begin{table*}
\centering
\caption{Top good attitude Facebook posts. These seem more puzzling than the
corresponding tweets, especially the 4th ranked post.}
\label{goodfb}
\begin{tabularx}{\textwidth}{|l|X|}
\hline
Score & Post \\
\hline
1.8313 & ought to be cast in the next big zombie movie as an extra. Because she feels like the living dead must feel after a night of heavy brain-eating. Is that their equivalent of drinking? Explains why they want to do it all the time. \\
1.803 & Out of the Main Event... my 11k didn't make it far into Day 2.  I've been playing cash games non-stop for the last two days though.  Probably going to play again today... Now that I'm out, planning to come to *PROPNAME*'s wedding  \\
1.7976 & Prayers needed for 12yr old boy who, after a serious car crash, is now in a critical coma. Pls. change your status for 1 hr so more people can become aware and add to the prayers. We would do it for your son or relative, pls. do this for somebody else's son. Thank you. Pls copy this and paste..  \\
1.6902 & Dear car battery: Why are you dead again?  What am I going to do with you?? (No, really--what am I supposed to do with it?)  You make me sad. ://  \\
1.6326 & has walked a thousand miles with grace so far. I could walk like this forever...Jesus how can you be so good to me?? May you be praised in every step I take. Always \\
1.5979 & I believe it all is coming to an end. Oh well, I guess we're gonna pretend  . Ever feel like a five-year-old who is content to cover his//her eyes to feel safe and secure? If I can't see you, you can't see me..lol..I want us all to stop ignoring the important things. They won't go away if you ignore them. I promise.  \\
1.5319 & is impressed that the ads on the side of FB are getting relevant without being offensively so...finally.  I've got an ad for a book I like, a band I like, and a movie by an animator I like.  Yay!  \\
1.5319 & RUDY IS COMING AGAIN!!! \\
1.5319 & Oh, and the 7th of January I'm going for my restricted license... again. Yay.  \\
1.5319 & After a slight rebuild... Betsy 2.0 lives to surf the net again!! \\
\hline
\end{tabularx}
\end{table*}

\begin{table*}
\centering
\caption{Top bad attitude tweets}
\label{badtweets}
\begin{tabularx}{\textwidth}{|l|X|}
\hline
Score & Post \\
\hline
-8.60 & Well we're getting a fair amount of the sunny stuff but if you're out at 2 its probably hard to see. R u waitin for a sunrise? \\
-8.60 & Cholos r neat,refuse 2 let the style die.where they still get cholo polos at?How does Nike n Reebok know 2 make the shoes just cholo right? \\
-8.55 & I CAN READ I JUST SEEN U SAY U WAS PUTIN HIM 2 SLEEP IM GON LEAVE U ALONE FOR RIGHT NOW CUZ I FEEL U MAD AT ME \\
-8.44 & well 1. you should try and pee at the latest 30 mins -1 hr after sex that helps get rid of most of it then you know just proper \\
-8.31 & you should be greatful of the sun... When you come to Dublin with Miley you will want some sun... We never get sun.. Just rain \\
-8.24 & So I'm just sitting at home on my own watching rubbish TV...is it wrong to go to bed when it's still slightly light outside?Or am I just old \\
-8.24 & Yeah, seriously. Then I just tried again and everything was fine. Still feel like I've been put in my place though. :) \\
-8.13 & user15380 I'm probably gonna get it onthe PS3 when I get paid Wednesday. I can even put my pic inthe action. Do it. I'll lend u my PSeye. \\
-8.02 & take a picture of me =P that way u can put it ina frame by your bed and u can smile at me b4 bed everynite, x.x i sound concited \\
-8.00 & I'm throwing out the first pitch at the Twins/Sox game tonight! What should I throw: my change, fastball, or hook? \\
\hline
\end{tabularx}
\end{table*}

\begin{table*}
\centering
\caption{Top bad attitude Facebook posts}
\label{badfb}
\begin{tabularx}{\textwidth}{|l|X|}
\hline
Score & Post \\
\hline
-2.4469 & [13 hours in vcpa today... it's going to be a long day..!] \\
-2.4287 & ...when will the GSS end? is BOSE headphone part of the GSS? wanna get one with my pay...when i get my pay...    \\
-2.4202 & I declare myself a mercenary. Have a job? I'll get it done. \\
-2.4057 & During a recent password audit, it was found that a blonde was using the following password: MickeyMinniePlutoHueyLouieDeweyDonaldGoofySacramento. When asked why such a long password, she said she was told that it had to be at least 8 characters long and include at least one capital.  \\
-2.3984 & finished Pride \& Prejudice \& Zombies instead of watching Ghost in the Shell.  I'm classy.  Also, a giant geek. :)  \\
-2.3984 & is a Fool but also a Dreamer \\
-2.3875 & will be completely honest for the next 24 hours....you can ask me one question (only in my inbox)....any question no matter how crazy or wrong it is....but you can only ask me one question!!!! you will have my full honesty....but i dare you to put this as your status too and see what questions you get !!! :) x This should be interesting. \\
-2.3221 & needs to get hair recolored. ahhhhhhhhhhhh \\
-2.3221 & I wanna get two bunnies....They are sooo cute.... \\
\hline
\end{tabularx}
\end{table*}

\begin{table*}
\centering
\caption{Top active attitude tweets. It is striking how many of these
tweets focus on the issue of getting up in the morning.}
\label{activetweets}
\begin{tabularx}{\textwidth}{|l|X|}
\hline
Score & Post \\
\hline
7.38 & I'm goooood!!! Might go to bed soon with a book! :D early night seeing as I need to get up early tomorrow! :'( xxxx \\
6.88 & Yes, well perhaps the very short preface of mine. The rest of the book is just pages of "ouch" with accompanying drawings. \\
6.88 & I just have it lying around so you may as well have it. About Offspring, have you seen their new video? Not very good :( \\
6.82 & OKAY! I need to get outta bed now to go to eat before I die of um this nua-ness. So much to do today! And I need a Fuji ISO 400 film. \\
6.65 & Well, I'm already up, damned world. But I may be as useless as a chimp who can't throw bananas...or some better metaphor \\
6.39 & That's right, I almost forgot. well, have a good day. 4x10's aren't so bad, especially when you have a 3 day weekend ;) \\
6.23 & you could ask them if you could get it switched, call them. But you will prob have to get a new one. Need help w/ new name? \\
6.14 & OK, I seriously may need someone to start MAKING my ass get up in the mornings....wish I could make Charlie get me up!!!! LOL... \\
\hline
\end{tabularx}
\end{table*}

\begin{table*}
\centering
\caption{Top active attitude Facebook posts. In contrast to the tweets,
these posts tend to focus on exercise.}
\label{activefb}
\begin{tabularx}{\textwidth}{|l|X|}
\hline
Score & Post \\
\hline
1.9414 & just received an extra 360 kb of storage for his computer in the form of a 5 1//4   floppy from the guy near me who's retiring.  It's going to come in handy for that extra 20 seconds of music that won't fit on my hard drive.  \\
1.8438 & [13 hours in vcpa today... it's going to be a long day..!] \\
1.8311 & long day full of yays and mehs... \\
1.807 & Fucking CARBON MONOXIDE alarm's going off repeatedly again tonight! Opened the windows \& garage doors; I have nowhere else I can go \& I need sleep, so I'm gonna hafta just hope the ventilation is enough cos I don't know what else to do. Hope I don't end up in the hospital over this shit--I'm sick of the damn hospital! (Well...bein' dead would [debatably] suck too. But I wouldn't be self-aware to know about it, lol.)  \\
1.7347 & thinks people should either do something or shut up and deal!  We were the (pretty much) first country to have a peaceful transfer of power between political parties.  Let's go back to those times, shall we?  If you're not gonna do anything about what you don't like, then please, shut up and leave everyone else alone. kthxbi.  \\
1.6869 & going for a long run hit me up later \\
1.6844 & gym=the hurt \\
1.6707 & In my shoes, just to see, what it's like to be me, I'll be you, let's trade shoes Just to see what it'd be like to feel your pain, you feel mine, go inside each others minds Just to see what we'd find, look at shit through each others eyes  \\
1.6683 & Most people never run far enough on their first wind to find out they've got a second \\
1.6437 & Hurt my leg today. Might have a stress fracture not sure yet. I can walk on my leg but it HURTS lol anyways cant wait to see the volley ball game! The Halloween party is still on tonight at my house. I live like 10 mins away from chief leschi so transportation shouldnt be that big of a problem \\
\hline
\end{tabularx}
\end{table*}

\begin{table*}
\centering
\caption{Top passive attitude tweets}
\label{passivetweets}
\begin{tabularx}{\textwidth}{|l|X|}
\hline
Score & Post \\
\hline
-6.93 & it's hard to believe that's the same girl that was rockin flour all over her face last year!!! :-D \\
-6,76 & Postcard from Paradise.-NL-You know how hard it is staying out of the sun here. Eclipse rehearsals on a week -URL- \\
-6.61 & The dog got 2nd prize at a talent thing at local pet shop - only 2 entrants :) so wasn't hard :) \\
-6.48 & aww man who you tellin...I tried hard to talk myself outta starting, now I'm tryna talk myself out of stopping...smh...lol \\
-6.42 & ysl is the shit vavy user3985 nigga i found all the hard yeve saint laurent shit the got shit for the low lol belts only 275 cop out boy! \\
-6.37 & Grindin hard all day true style of a hustler...not 2 many females out here doin it like that....Well I must say I am one and will remain \\
-6.27 & invader zim : if you do not know it, we must, must, must provide you with data ... \\
-6.17 & Igor will start the STEPS soon!!!  Be ready!!!!  Do not miss out this time!  Too many fans said they regret not getting twisted! No regrets! \\
-6.16 & user950 hope you do that quickly! come back soon! \\
-6.12 & Visited bro in hospital. Got a roommate @ the last minute. 80+ year old man who made sure to tell the nurse he needs a commode. Poor bro. \\
\hline
\end{tabularx}
\end{table*}

\begin{table*}
\centering
\caption{Top passive attitude Facebook posts. It is striking that these
are all short, that is they use only a few words but these
are strongly passive.}
\label{passivefb}
\begin{tabularx}{\textwidth}{|l|X|}
\hline
Score & Post \\
\hline
-2.8905 & scored another interview!!!!!! \\
-2.8905 & earned another freckle or two from running the *PROPNAME* Marathon today!! \\
-2.8905 & accidentally rescued another animal from T\&D's... \\
-2.8905 & needs another weekend \\
-2.2181 & crowd surfed at *PROPNAME* and *PROPNAME*! =D \\
-2.1696 & is going to the lone star ruby conf. in August! \\
-2.0909 & And there goes another one!!!! \\
-2.0467 & Today is another day. You dont get better by making a right choice once. You get better by making the right choice... everyday... over and over again... \\
-1.9894 & is drowning in a sea of irrelevance. \\
-1.7607 & is seeing *PROPNAME* this weekend. :)  We're going to the Shedd Aquarium and doing other fun stuffs. <3 \\
\hline
\end{tabularx}
\end{table*}

\begin{table*}
\centering
\caption{Top strong attitude tweets. It is striking how much these
focus on issues related to sleep.}
\label{strongtweets}
\begin{tabularx}{\textwidth}{|l|X|}
\hline
Score & Post \\
\hline
8.58 & Good Night Good Morning, hope you had a peaceful sleep and some well needed rest \\
8.18 & Good Morning Tweeps!  Morning user7234 - get brekky yet?  Welcome back user7228 - get some sleep, good to have you back! \\
8.16 & Sleep is very good. I went into a deep sleep for the first time in a month. \\
7.28 & yeah I do get a headache when I sleep more than 8 hours and when I sleep when I'm angry 2.....and on most Thursdays \\
7.20 & Good morning! How did you sleep? Did you shake off the undead? \\
7.12 & yep @user14449 me too I feel like hols have begun! Hehe I wish! Yep jus the usual - sleep and catch ups! We must catch up with Nat as well! xo \\
6.66 & "are you watching me sleep walter?" "yes your beautiful when you sleep" "okay watch me sleep again" "kay shut your eyes, OMG its BEAUTIFUL!" \\
6.63 & I must sleep again. Good night! xD \\
6.62 & good morning or afternoon, or night lol its morning here,  :) sleepy gotta get more then 4 hours sleep a night lol how r u? \\
6.52 & Describe to me what a good night of sleep is like. I really can't recall... \\
\hline
\end{tabularx}
\end{table*}

\begin{table*}
\centering
\caption{Top strong attitude Facebook posts}
\label{strongfb}
\begin{tabularx}{\textwidth}{|l|X|}
\hline
Score & Post \\
\hline
3.1886 & amber decided it would be fun to steal an egg and smash it on the carpet ): \\
2.0967 & wants to live where soul meets body, and let the sun wrap its arms around her, and bathe her skin in water cool and cleansing, and feel, feel what it's like to be new.  \\
2.0482 & hmm, can't wait to try out this   synthetic fermented egg   lure...  \\
1.8823 & is reading   Act like a lady, think like a man   on her lunch...understanding the male psyche is the first step to taking over the world :oP  \\
1.8577 & is drowning in a sea of irrelevance. \\
1.7469 & just doesn't feel like working... how about sleep!? :) \\
1.7082 & just heard a commercial on the radio about some pill to lower cholesterol. They actually said   Do you have high cholesterol due to poor diet or lack of exercise? Then you need to try this    No, actually you need to try improving your diet and exercising a bit first.  Just a crazy thought I had...  \\
1.6636 & is quite liking girls aloud \\
1.6636 & mmmm beeen feeling quite happpy =)))) \\
1.6468 & Nothing quite like going to the park and seeing a vulture munch on a gutted squirrel, turn to look at you, and precede to empty its bowels with a loud *squirt*  \\
\hline
\end{tabularx}
\end{table*}

\begin{table*}
\centering
\caption{Top weak attitude tweets}
\label{weaktweets}
\begin{tabularx}{\textwidth}{|l|X|}
\hline
Score & Post \\
\hline
-10.61 & yeah poor little baby rosco will have to be in the way back all by himself u need to light a fire under my butt about my blog lol \\
-9.39 & Pt 2: Face Place for facials-Tony and Paul are the best hands down in the world \& I have been 2 the best places all over the world 2 compare \\
-9.35 & I need u all 2 send positive weather thoughts my way. Fundraising BBQ 2nite and its been tipping it down all day here. Typical. \\
-8.22 & user6068 user6069 OK guys... just checked nkotb.com ... all the threads are about not being able to book. So WHO is on the boat then??? \\
-7.89 & Le sigh. Thing is, I have a lot of planned posts right up here in my head. But then this is just so much easier. Wahhh. \\
-7.77 & Fight Club is one of my favorite movies; so much so, it got me to read. And I don't read much beyond a few pages of books \\
-7.52 & All he done was show the bad points about Michael and try to make him out as a dick. I hate Martin Bashir so much. \\
-7.45 & Sometimes what we need and are looking for is right under our nose all along. \\
-7.31 & I want to win something in this contest since I'm only 12 (no income here, only pocket money) and need iTunes money to buy \#simplytweet! \\
-7.30 & You're obviously not thinking about having fun all day. Sometimes you just need a smile on your face and everything works out. \\
\hline
\end{tabularx}
\end{table*}

\begin{table*}
\centering
\caption{Top weak attitude Facebook posts. It is striking how short they
are; again, few words, intensely weak.}
\label{weakfb}
\begin{tabularx}{\textwidth}{|l|X|}
\hline
Score & Post \\
\hline
-2.44 & my ankle hurts.  stupid dishwasher door \\
-2.2181 & 6k+car break down+friday the 13th=me feeling superstious. 6k=fail=me tired \\
-2.2181 & will be in knoxville for break. >.<' \\
-2.2181 & working until 1:00pm...really needs a break \\
-2.2181 & IT'S CHRISTMAS BREAK!!!!!!!!!! \\
-2.1072 & Kindergarten report cards \\
-2.1072 & is making his report. \\
-2.1072 & lingaw ako report sa 109. yey :p \\
-2.031 & Today is my 2 month birthday! \\
-1.9617 & work 12-7... \\
\hline
\end{tabularx}
\end{table*}

The results from the Osgood model are not as compelling as the results
for detecting emotions. We suspect that this is because tweets and
Facebook posts are only attitudinal in quite diffuse ways, that is
they reflect a general attitude to the world at the moment of
authoring and so conveyed in words whose weightings are quite small.

\section{Discussion and conclusions}

We have considered the problem of inferring multiple aspects of
human affective state from the perspective of a robot which
(over)hears the conversation of humans in its vicinity.
For inferring rapidly changing components of affective space,
such as emotions, computations must be done from small amounts 
of text and in short time frames. For components that change more
slowly, greater amounts of text and longer time frames are
plausible.

A robot must understand the emotional state of humans with which
it interacts to complete the affective loop, which seems to
be essential to creating truly social interactions. Mood, in turn,
must be understood because of its modulating effect on emotion.
The ability to detect wishes, desires, and preferences is the first
step towards a robot that can act without explicit commands in
ways that the humans it serves want, reducing the cognitive burden
on those humans.
The ability to infer attitudes and personality are useful for a
robot to winnow, in advance, the likely repertoire of decisions and 
actions a human might take in particular situations, and so allow
the robot to precompute likely responses it might be called upon to
make.

We use tweets, headlines, and Facebook posts as surrogates for conversations,
and essays and vlog transcripts as surrogates for what a robot might 
accumulate about
a human in the course of multiple interactions. These documents
are extremely ungrammatical, containing multiple abbreviations,
misspellings, and jargon. Some of these would remain in conversational
snippets, but some would not.

We do not have ground truth for the emotions, wishes, desires, and
attitudes of the authors of each document. We appeal to face
validation by showing those documents that rank most highly for
each of the properties considered.
In general, it is clear that the high-ranked documents do exhibit the
relevant property but, of course, this does not mean that other
documents that also exhibit the property have not been missed.
For personality, we use datasets for which ground truth is known.
Even humans typically achieve no better than
80\% accuracy for personality trait prediction.
Hence, our algorithm's performance, achieving prediction
accuracies mostly in the 70\% range, is strong.

The surprising conclusion of our analysis is that computational models
perform usably well, even from quite small amounts of text, and with
relatively simple analysis. It is therefore plausible that robots
could build models of the affective state of the humans they
encounter using practical amounts of computation time and resources.

\appendix
\section*{Avoiding recomputing the SVD}

If $\hat{A}$ is the matrix with extra (test) rows added, then
\[
\hat{V} \;\approx\; \hat{A} {U^{+}}' S^{-1}
\]
where $U^+$ is the Moore-Penrose inverse of $U$ (which isn't square,
let alone invertible) given by $U^+ = (U' U)^{-1} U'$. $\hat{V}$
is $m \times k$, $\hat{A}$ is $m \times (n+q)$, ${U^{+}}'$
is $(n+q) \times k$, and $S^{-1}$ is $k \times k$.

\bibliographystyle{plain}
\bibliography{affect}
\end{document}